%% file: main.tex
\documentclass[conference]{IEEEtran}
\IEEEoverridecommandlockouts

\input{com/header}

\begin{document}
    \title{\OurSolution: Abnormal Traffic Flow Detection\\ Using Siamese Networks}

    \author{\IEEEauthorblockN{Sepehr Sabour, Sanjeev Rao and Majid Ghaderi}
    \IEEEauthorblockA{Department of Computer Science, University of Calgary\\
    \{\tt sepehr.sabour, sanjeev.rao, mghaderi\}@ucalgary.ca }
    {\footnotesize \thanks{This work was supported by Wedge Networks Inc., Alberta Innovates and Natural Sciences and Engineering Research Council of Canada.}}
    }

    \maketitle

    \input{./tex/abs.tex}
    \input{./tex/introduction.tex}
    \input{./tex/recent_works.tex}

    \input{./tex/system.tex}
    \input{./tex/methodology.tex}

    \input{./tex/dataset.tex}

    \input{./tex/evaluation.tex}
    \input{./tex/conclusion.tex}

    \balance

    \bibliographystyle{IEEEtran}
    \bibliography{IEEEabrv,bib/main}

\end{document}

%% file: com/header.tex
\usepackage{amsthm}
\usepackage{amssymb}
\usepackage{amsmath}
\usepackage{graphicx}
\graphicspath{{figs/}}
\usepackage[font=small,belowskip=-5pt]{caption}
\usepackage[labelformat=simple,font=scriptsize]{subcaption}

\usepackage[dvipsnames]{xcolor}
\usepackage{cite}
\usepackage{hyperref}
\hypersetup{nolinks=true}
\usepackage{algorithm}
\usepackage{algorithmic}
\usepackage{lastpage}
\usepackage{setspace}

\usepackage{textcomp}
\usepackage{array}
\usepackage{booktabs}
\usepackage{amsfonts}
\usepackage{tabularx}
\usepackage{graphicx}
\usepackage[export]{adjustbox}
\usepackage{balance}
\usepackage{nicefrac}
\usepackage[normalem]{ulem}
\usepackage{siunitx}
\usepackage{relsize}
\usepackage{multirow}

\newcommand{\eg}{\textit{e.g. }}

\newcommand{\eqend}{\,.}

\newcommand{\OurSolution}{\textsmaller{\textsf{DeepFlow}}}
\newcommand{\Section}[1]{\textit{Section~\ref{#1}}}
\newcommand{\Fig}[1]{Fig.~\ref{#1}}
\newcommand{\Table}[1]{Table~\ref{#1}}
\newcommand{\Exp}[1]{Exp.~\ref{#1}}

\newcommand{\cat}[1]{\medskip\noindent\textbf{#1.}}
\newcommand{\subcat}[1]{\noindent\textbf{#1.}}

%% file: tex/abs.tex
\begin{abstract}
    Nowadays, many cities are equipped with surveillance systems and traffic control centers to monitor vehicular traffic for road safety and efficiency.
    The monitoring process is mostly done manually which is inefficient and expensive.
    In recent years, several data-driven solutions have been proposed in the literature to automatically analyze traffic flow data using machine learning techniques.
    However, existing solutions require large and comprehensive datasets for training which are not readily available, thus limiting their application.
    In this paper, we develop a traffic anomaly detection system, referred to as \OurSolution, based on Siamese neural networks, which are suitable in scenarios where only small datasets are available for training.
    Our model can detect abnormal traffic flows by analyzing the trajectory data collected from the vehicles in a fleet.
    To evaluate \OurSolution, we use realistic vehicular traffic simulations in SUMO.
    Our results show that \OurSolution\ detects abnormal traffic patterns with an $F_1$ score of $78\%$, while outperforming
    other existing approaches including: Dynamic Time Warping~(DTW), Global Alignment Kernels~(GAK), and iForest.

\end{abstract}


%% file: tex/introduction.tex
\section{Introduction}
\label{I}

\subcat{Motivation}
Driving safety continues to be a challenging problem in city management.
Based on a road safety plan published by \textit{Canadian Council of Motor Transport Administrators}, about $2000$ people are killed, and $165,000$ are injured in car accidents annually in Canada \cite{R10}.
Distracted driving leads to unusual actions such as sudden accelerations, decelerations, and lane changes that other vehicles cannot predict, which can result in collisions.
%
Technology is expected to play a significant role in road safety to lead the transportation system toward zero fatal accidents~\cite{R10}.
An Intelligent Transportation System (ITS) makes use of technologies such as vehicular networks, cloud computing, and artificial intelligence to solve traffic flow problems.
For example, ITSs employ Vehicle to Everything (V2X) communications~\cite{ITS} to collect information related to vehicles, pedestrians, and road conditions.
This data can be used to analyze traffic flows and driver behavior in order to detect abnormal driving patterns.
Detecting abnormal behavior in vehicular traffic, apart
from improving transportation safety, significantly impacts evaluation of driving skills, including that of autonomous vehicles.
For instance, insurance companies can base their premiums on one's driving behavior.
Also, understanding the misbehavior of autonomous vehicles can help analyze the risks involved in detaching human drivers from vehicles.

The emergence of Machine Learning (ML) has paved the way for more efficient solutions for many of the challenges faced in various fields such as
fraud detection, cyberattack prevention and anomaly detection.
ML solutions can help reduce the system dependence on human-in-the-loop processes in order to boost performance and reduce cost.
Traffic management in cities can benefit from this idea too.
A modern traffic control center is equipped with several display devices to monitor daily traffic flows in a city.
The operators in these centers continuously watch for abnormal events on the roads to make sure traffic flows are steady and safe.
In addition, these centers provide helpful information to the emergency units in case of accidents.
However, manually checking traffic cameras in a city is inefficient and expensive. An automated system to check for traffic anomalies is essential for continuous and real-time analysis of vehicular traffic.
%

Existing solutions such as~\cite{R1,R2,R4,R6,R19} use a dataset of normal driving patterns, and mark any unseen pattern as an anomaly.
However, several factors like weather condition, road side constructions and traffic load
can change the behavior of vehicles.
Therefore, a huge dataset of normal patterns is required, which is not easy to acquire.
Other solutions rely on finding outliers in traffic flows~\cite{R3,R5}.
These approaches are based on two assumptions. First, the driving data (\eg trajectory, speed, and acceleration)
of abnormal vehicles diverges from normal ones. Second, normal vehicles form the majority of a given traffic flow.
However, the behavior of normal vehicles in a fleet varies over time,
which makes it challenging for such approaches to distinguish between normal and abnormal patterns.
For example, drivers may usually drive within $\pm 10\%$ of the speed limit and still be considered to be driving normally.

%
\vspace{-0.2cm}
\cat{Our Work}
In this work, we introduce \OurSolution, an anomaly detection system which detects abnormal traffic flows by \textit{analyzing vehicle trajectories in a fleet}.
We use a small set of normal cases to train our model, and test it with a dataset containing previously unseen patterns.
We show that \OurSolution\ can address the challenges faced by existing approaches.
The model learns the similarity between vehicles, assigns a score based on that, and classifies flows based on this similarity score.
Our realistic experiments show \OurSolution\ is effective \textit{even when a comprehensive dataset is not available for training}.

The main idea behind our solution is that vehicles in a normal traffic flow have similar trajectory data,
so the presence of any abnormal cases can be detected by measuring the average similarity between vehicles.
For example, when a car drives with a higher speed and acceleration than other ones or abnormally changes lanes,
its trajectory data is different than others.
To measure this similarity, we use neural networks to compress the data-series from each vehicle into a latent vector,
and we measure the distances between them.

Our main contributions in this paper are:
\begin{itemize}
    \item We present the design of \OurSolution,
    a traffic anomaly detection system, and describe its various components including data collection, anomaly detection and application.
    \item We design, implement and evaluate a semi-supervised Siamese network to measure the abnormality score of traffic flows within a fleet of vehicles.
    \item We use realistic traffic simulations in SUMO to obtain datasets for training and testing \OurSolution.
    Our results show that \OurSolution\ detects abnormal traffic patterns with $78\%$ $F_{1}$ score and outperforms
    other existing approaches including: Dynamic Time Warping~(DTW), Fast Global Alignment Kernels~(GAK) and iForest.
\end{itemize}

\cat{Paper Organization}
We review recent work on traffic anomaly detection in \Section{II}.
The design of \OurSolution\ is discussed in \Section{III}.
The anomaly detection engine used in \OurSolution\ is presented in \Section{IV}.
In \Section{V}, we describe the simulation process and our datasets.
Evaluation results are presented in \Section{VI}, while \Section{VII} concludes the paper.

%% file: tex/recent_works.tex
\section{Related Works}
\label{II}

In the following, we briefly review several representative papers on traffic anomaly detection that are most relevant to our work.
We categorize available works into \textbf{history-based} and \textbf{outlier-detection} approaches.

\cat{History-Based Approaches} In this approach, the behavioral history of vehicles is used to check for the presence of anomalies in real-time.
For instance, SafeDrive~\cite{R1} is a driving anomaly detection approach that uses historical data to generate a state
graph in which states represent the value (or its range) of sensor data, and weighted edges show the likelihood of transitions between the states.
At the beginning of the driving path, the vehicle is in the starting state, and by receiving the real-time driving events,
the state changes.
One can measure the driver's anomaly score by aggregating the weight of the traveled edges.
Similarly, authors in~\cite{R2} employ a graph-based approach and reinforcement learning techniques to detect abnormal trajectories.

Neural networks such as autoencoders~\cite{Goodfellow} and LSTMs~\cite{LSTM} have also been used for history-based anomaly detection.
For example, a technique for detecting anomalies using autoencoders is proposed in~\cite{R4}.
Similarly, authors in~\cite{R6} propose an anomaly management system which uses autoencoders to find abnormal drivers in a collaborating transportation system.
Driving behavior prediction is another approach to identify anomalies.
To give an example, authors in~\cite{R19} apply two different solutions containing a recurrent neural network (RNN) and
a long short-term memory (LSTM) to predict driver's actions, and mark behaviors that are varying from the predicted ones.

\cat{Outlier-Detection Approaches}
These approaches compare the behavior of vehicles and mark outliers.
For instance, in~\cite{R3}, three ML algorithms, namely Support Vector Machine~(SVM), Isolation Forest~(iForest), and K-Nearest Neighbors~(K-NN),
are used to detect outlier drivers.
Also, the authors of~\cite{R5} present a reckless driver detection framework which uses vehicular collaboration
to collect data and then apply support vector machine (SVM) and decision-tree models to measure every vehicle's driving performance.

In general, history-based approaches require a large dataset. On the other hand, SVM and K-NN are supervised approaches, which need a dataset containing abnormal cases for training.
However, \OurSolution\ can operate with a small dataset, and because it is a semi-supervised method, no datasets containing abnormal cases are required for training.

%% file: tex/system.tex
\begin{figure}[t]
    \centering
    \includegraphics[width=8cm]{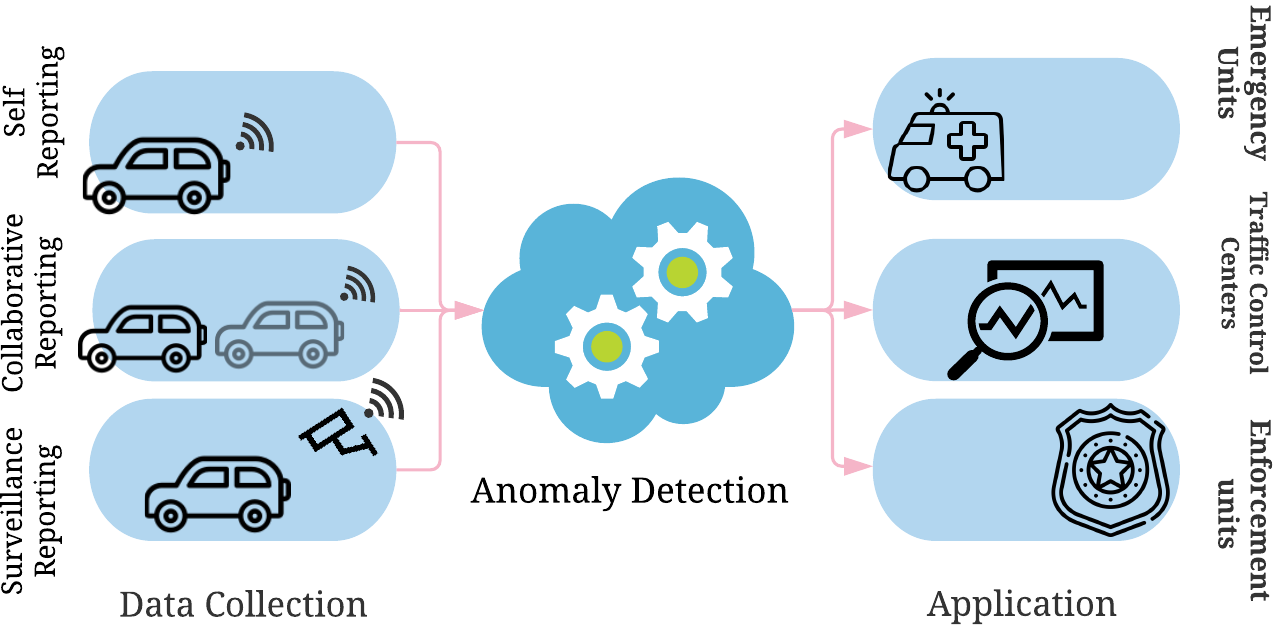}
    \caption{High-level design of \OurSolution. \vspace{-0.5cm}}
    \label{fig:system}
\end{figure}

\section{\OurSolution\ Design}
\label{III}

\Fig{fig:system} shows the high-level design of \OurSolution. As shown in the figure, \OurSolution\ contains three components including data collection, anomaly detection and application.
In the following, we describe each of these components.

\subsection{Data Collection}
We feed \OurSolution\ with time-series data collected from a group of vehicles;
this contains information such as speed, location and steering angle.
The data can be gathered using one or a combination of the following approaches:

\cat{Self Reporting}
Most modern vehicles are equipped with sensors and onboard communication devices due to the industry-wide push
towards more automated vehicles.
These sensors measure the speed, location, and gap between a vehicle and the surrounding objects.
Vehicles can report their data to the server for anomaly detection.
The self-reported data to the server is more accurate than other approaches.
Nevertheless, it demands that all vehicles have the required hardware.
Also, an abnormal vehicle can manipulate or avoid sending data to stay hidden from the anomaly detection system.

\cat{Collaborative Reporting}
In this approach, each vehicle measures the state of the adjacent vehicles and reports it to the server.
The information inferred by the other neighbors may be inaccurate;
however, participating vehicles can reach a consensus on its validity using vehicle to vehicle (V2V) communication.
Therefore, the system understands the correct state of the traffic flow even if an adversarial vehicle blocks or changes the data.
However, like any other wireless network solution,  jamming and corrupting the communication signals is possible.

\begin{figure}[t]
    \centering
    \includegraphics[width=8cm]{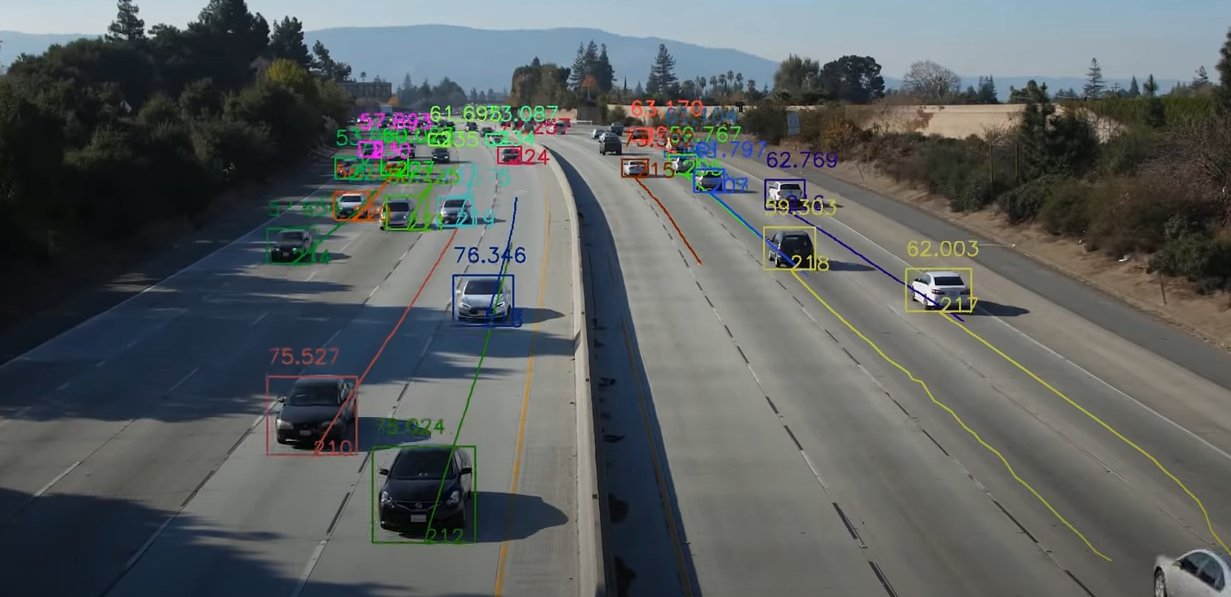}
    \caption{Tracking vehicles using a surveillance camera~\cite{R14,R15}.}
    \vspace{-0.3cm}
    \label{fig:tracking}
\end{figure}

\cat{Surveillance Reporting}
Surveillance systems such as traffic cameras and road sensors can capture information like the speed and location of the vehicles.
Image processing techniques can be used to process the collected video feeds and extract traffic flow information.
For example, \Fig{fig:tracking} shows a demo of such a vehicle tracking system~\cite{R14,R15}.
Despite the processing overhead of this approach, it does not require any hardware installed on the vehicles.
Also, since there is a direct connection between the surveillance devices and the cloud server, it has a lower security risk.

\subsection{Anomaly Detection}
\begin{figure}[t]
    \centering
    \includegraphics[width=9cm]{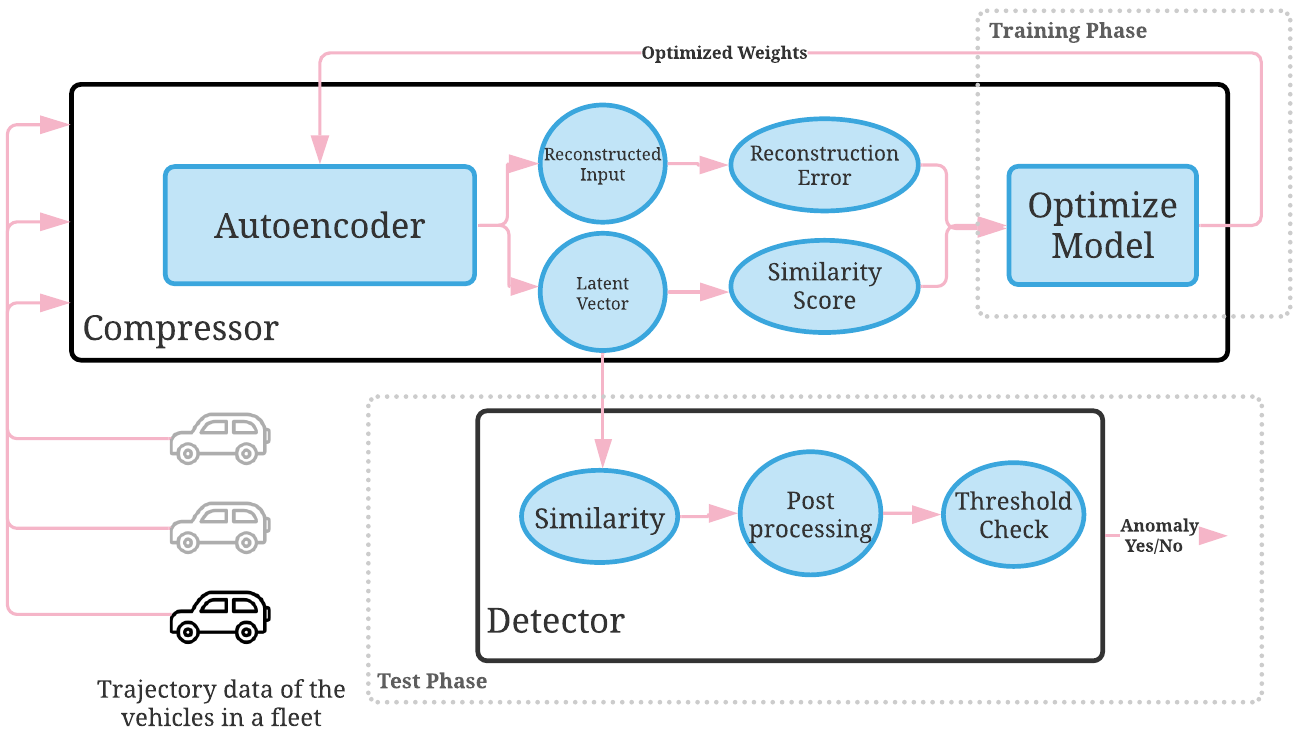}
    \caption{Structure of \OurSolution's Siamese network. \vspace{-0.5cm}}

    \label{fig:arch}
\end{figure}

At the core of \OurSolution\ is a Siamese network, as depicted in Fig.~\ref{fig:arch}.
A Siamese network contains two identical neural networks with the same weights.
This model is used to measure the similarity of two vectors by feeding them to the twin networks and comparing their outputs~\cite{siamese}.
Siamese networks are very useful in applications where no comprehensive dataset exists for training.
The anomaly detection component applies a pre-trained machine learning model on each vehicle's trajectory data and outputs an anomaly score for it.
The detector can function on a cloud or an edge server.
For the sake of privacy requirements and decreasing the cost of storage requirements, the server can eliminate all the processed data after making
the required computations.
In \Section{IV}, we delve into the structure of the applied machine learning model in this component.

%

\subsection{Application}
The output of the anomaly detection can be used in multiple applications as described below.

\subcat{Traffic Flow Analysis}
By employing this system we can measure the average abnormality score of
traffic flows in streets, junctions and highways,
which can help optimize the vehicular movements in these areas.

\subcat{Enforcement of Traffic Laws}
Enforcement units make use of this system to detect aggressive and distracted drivers.
The detected abnormal drivers can be penalized based on their anomaly scores.

\subcat{Emergency Situation Detection}
Traffic accidents can affect the abnormality score of a flow, which makes
\OurSolution\ able to mark sudden changes in scores and notify nearby emergency units.

This paper mainly focuses on implementation and evaluation of the anomaly detection component.
Therefore, elaboration of the data collection and application components is out of the scope of this article.

%% file: tex/methodology.tex
\section{Anomaly Detection Engine}
\label{IV}

\OurSolution\ employs a semi-supervised machine learning approach based on a Siamese network to detect abnormal traffic patterns.
\Fig{fig:arch} shows the architecture of this network.
The input of the network is a set of trajectory data-series collected from vehicles driving in a fleet.
At the first step, the \textit{Compressor} converts the data from each vehicle to a latent vector.
Then, the \textit{Detector} measures the distance between the vectors,
specifies a similarity score for each one, and finally measures the abnormality score of the flow.
In the following sub-sections, we explain the architecture, objectives, and functionality of each component in detail.

\subsection{Compressor}
The compressor is intended to shrink the input into a compressed ``latent representation", which
is accomplished by using an autoencoder (AE).
An autoencoder is a type of artificial neural network trained to compress the input and then decompress it with minimal distortion.
These neural networks are composed of two parts, an \textbf{encoder} that imposes a bottleneck and compresses the data,
followed by a \textbf{decoder} which reconstructs the input from the compressed representation~\cite{Goodfellow}.

Usually, the only objective of an AE is to minimize the reconstruction error.
In our system, we use the Mean Squared Error (MSE) to measure the error.
Also, to simplify the learning process, we use a $\tanh$ function to limit the error value in a range between $0$ and $1$. Thus,
the reconstruction loss function in \OurSolution\ can be expressed as follows:
\begin{equation}
    \mathrm{RLoss}(Y,\widehat{Y}) = \tanh\Big( \frac{1}{n} \sum_{i=1}^{n}{(Y_i - \widehat{Y_i})}^{2} \Big),
\end{equation}

where $Y_i$ and $\widehat{Y_i}$ are the $i$th actual and the $i$th reconstructed values of the input vector with size $n$, respectively.

To use the latent representation for outlier detection, we consider the compressed representation of an outlier pattern to be different from the regular patterns.
In the training phase, we use a dataset containing only normal trajectories to train our model to maximize the similarity of latent vectors.
Hence, we add a second error function, similar to the first one, to measure the distance between latent spaces:
\begin{equation}
    \label{ex2}
    \mathrm{Sim}(L_{i},L_{j}) = \tanh\Big( \frac{1}{n} \sum_{k=1}^{n}{(L_i[k] - L_j[k]})^{2} \Big),
\end{equation}
where $L_i$ and $L_j$ are the latent vectors, created for vehicles $i$ and $j$, respectively.
In \OurSolution, we calculate the aggregated loss in each training iteration and optimize the neural networks to minimize this value.
%
We use the following expression to determine the aggregated loss:
\begin{equation}
    \begin{split}
        \label{loss}
        \mathrm{Loss} =
        & \frac{1}{m} \sum_{i=1}^{m} \mathrm{RLoss}(Y_{i},\widehat{Y}_{i})\\
        & + \frac{2\lambda}{m(m-1)}  \sum_{i=1}^{m}\sum_{j=i+1}^{m} \mathrm{Sim}(L_{i},L{j}),
    \end{split}
\end{equation}

where $Y_{i}$, $\widehat{Y}_{i}$ and $L_{i}$ are the input, reconstructed input, and latent representation of vehicle $i$, respectively.
Also, $m$ is the number of vehicles in the fleet, and $\lambda$ indicates the importance of reconstruction accuracy over the similarity score.

\begin{figure}[t]
    \centering
    \includegraphics[width=9cm]{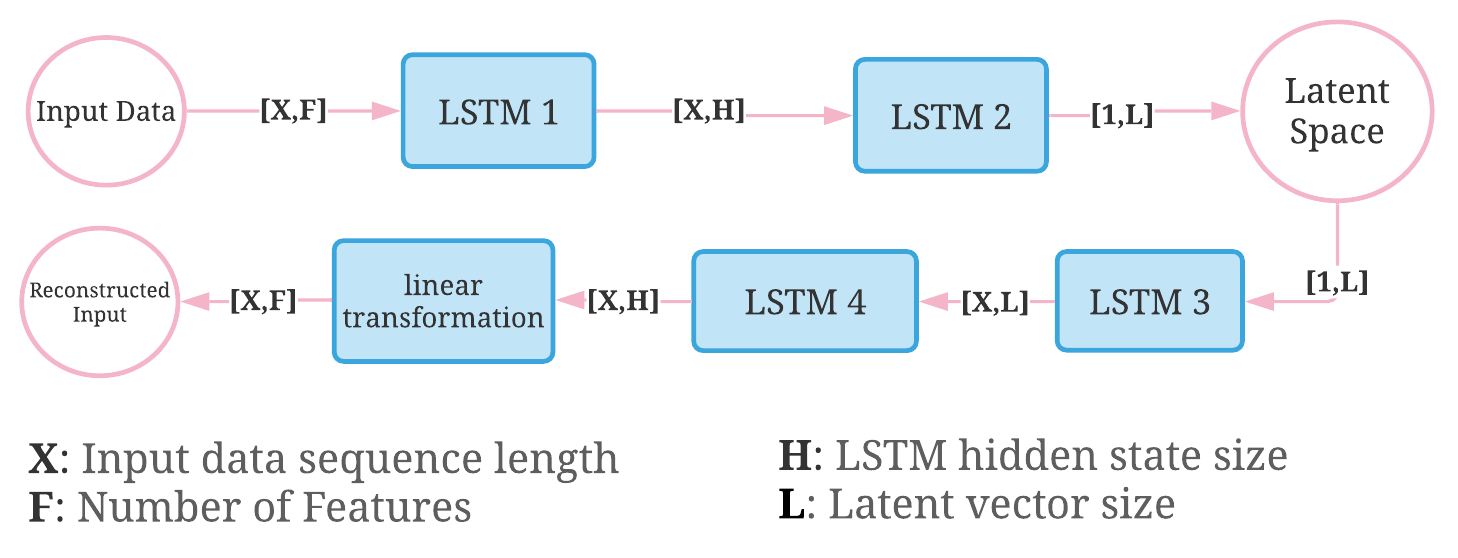}
    \caption{Structure of \OurSolution\ autoencoder. The arrows show the flow of data and indicate the input and output dimensions at each point in the network.
    \label{fig:model}  }

\end{figure}

As shown in \Fig{fig:model}, \OurSolution's autoencoder consists of four LSTMs and one linear neural network. 
LSTMs use a shared state between the nodes to remember the changes in the input over time.
In each step, a sequence of mathematical processes decides which part of the data should be remembered through the shared state and which part should be eliminated~\cite{LSTM}.
The first two networks in our autoencoder change the dimension of the input data and convert it to the latent vector.
Then, the next two LSTMs increase the size of the vector;
finally, the linear neural network unit reconstructs the input.
The compressor unit has two outputs: the latent representation and the reconstructed input.
In the training phase, we use both outputs to optimize the model using~\eqref{loss}.
However, we only use the first output in the testing phase and pass the latent vector to the detector unit.

\subsection{Detector}
After training the compressor, the system can be used to detect anomalies.
The Detector compares the latent space of each time-series data using an MSE function which is
similar to the loss function used in the training phase.
We use the following expression to calculate the abnormality score of traffic flows:
\begin{equation}
    \mathrm{AS} = 1 - \frac{1}{m(m-1)}  \sum_{i=1}^{m}\sum_{j=i+1}^{m}Sim(L_{i},L{j})
    \eqend
\end{equation}
After calculating the score, we compare it with a threshold to decide whether the traffic is normal or not.
A high score suggests low similarity between the behavior of the vehicles, which indicates an anomaly is present.
Therefore, finding an optimal threshold is critical.
In \Section{IV}, we study two approaches for finding a suitable threshold.

\subsection{Implementation}
The model is implemented in Python using PyTorch as explained below.

\cat{PyTorch}
To implement the model, we use \textit{Python3} programming language and \textit{PyTorch}~\cite{R16}.
\textit{PyTorch} is an open source deep learning library which facilitates building ML models by providing
basic machine learning modules such as LSTMs and linear neural networks.
We use the predefined models in PyTorch to construct the Siamese network~\footnote{Our source code is available at: https://github.com/pesehr/DeepFlow} in \OurSolution.

\cat{PyTorch Lightning}
This library~\cite{lightning} allows us to run neural network models on any hardware (CPU, GPU, TPU) with no changes required in the source code.
This feature is helpful as our development is done on a desktop computer, while model training is conducted on an Ubuntu Linux server with two V100 GPUs.

\cat{Weights \& Biases}
One of the main challenges in this work was finding the proper number of epochs for training our model.
This is important because, first, we try to find the minimal loss value which is important to prevent the model from over-fitting the training dataset.
Second, we should monitor the value of the two loss functions. At the beginning of the training phase, the value of the reconstruction error is significantly higher than the similarity error.
After several iterations, the model learns to reduce the first loss value and proceeds to minimize the second one.
Therefore, we need to monitor these values to stop the training after a suitable number of iterations.
For this, we used \textit{Weights \& Biases (WandB)}\cite{R18} to monitor the training process.

%% file: tex/dataset.tex
\begin{figure}[t]
    \centering
    \hspace{-1cm}
    \fbox{\includegraphics[width=6.5cm]{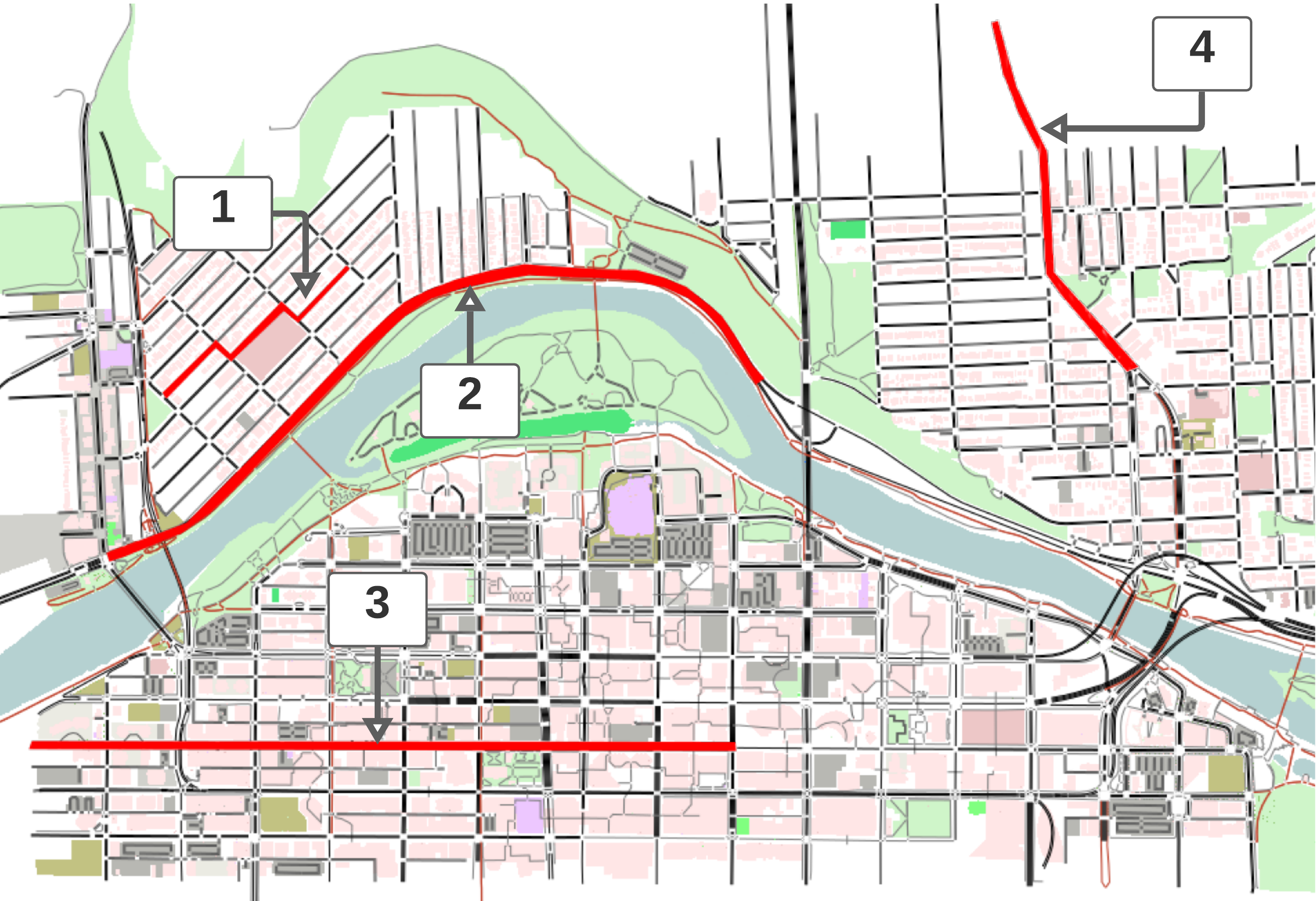}}
    \caption{Simulated City of Calgary traffic flow using SUMO. The highlighted paths are used for generating the test dataset.}\label{fig:calgary}
\end{figure}

\section{Traffic Flow Dataset}
\label{V}

The input of \OurSolution\ is trajectory data collected from a group of vehicles in a fleet.
Due to the difficulty in obtaining a dataset from a group of vehicles in an actual scenario, we generated the necessary datasets using simulation.
The datasets used in our evaluations are generated using the road traffic simulator software \textit{Simulation of Urban Mobility (SUMO)}~\cite{R11}, which is
an open-source traffic simulator capable of simulating different components involved in a traffic scenario such as roads, vehicles and pedestrians.

\subsection{Traffic Simulation}
In order to use this software, the following elements should be specified:

\begin{itemize}
    \item  \textbf{Network File:} The location and shape of each road, junction, and sidewalk.
    Also, the network file indicates the traffic rules such as direction, priority, and speed limit of each path.

    \item \textbf{Traffic Demand File:} Determines how many vehicles are in the system and describes their behavior.
    Also, each driver's arrival and departure time, and the path taken by them should be defined in the traffic demand file.
\end{itemize}

SUMO includes several tools to help with the simulation process.
We use the following tools in our work:

\cat{OSMWebWizard} A Python script implemented to work with \textit{OpenStreetMap}\cite{R13} which extracts network data from the actual street map.
In this work, we use it to simulate downtown Calgary and the surrounding regions.
We select this area due to the variety of available streets. \Fig{fig:calgary} shows the area of the city which is simulated.

\cat{TraCI} An interface implemented by SUMO to get data values from simulated vehicles and control their behavior.
It is available as a Python library and employs a TCP-based client-server architecture to access the simulator \cite{R12}.
We use \textit{TraCI} to manipulate the speed of the vehicles and create the scenarios that are used in our training dataset.
\begin{figure}[!htb]
    \hspace{-1.5cm}
    \begin{minipage}{0.6\linewidth}
        \includegraphics[width=\linewidth]{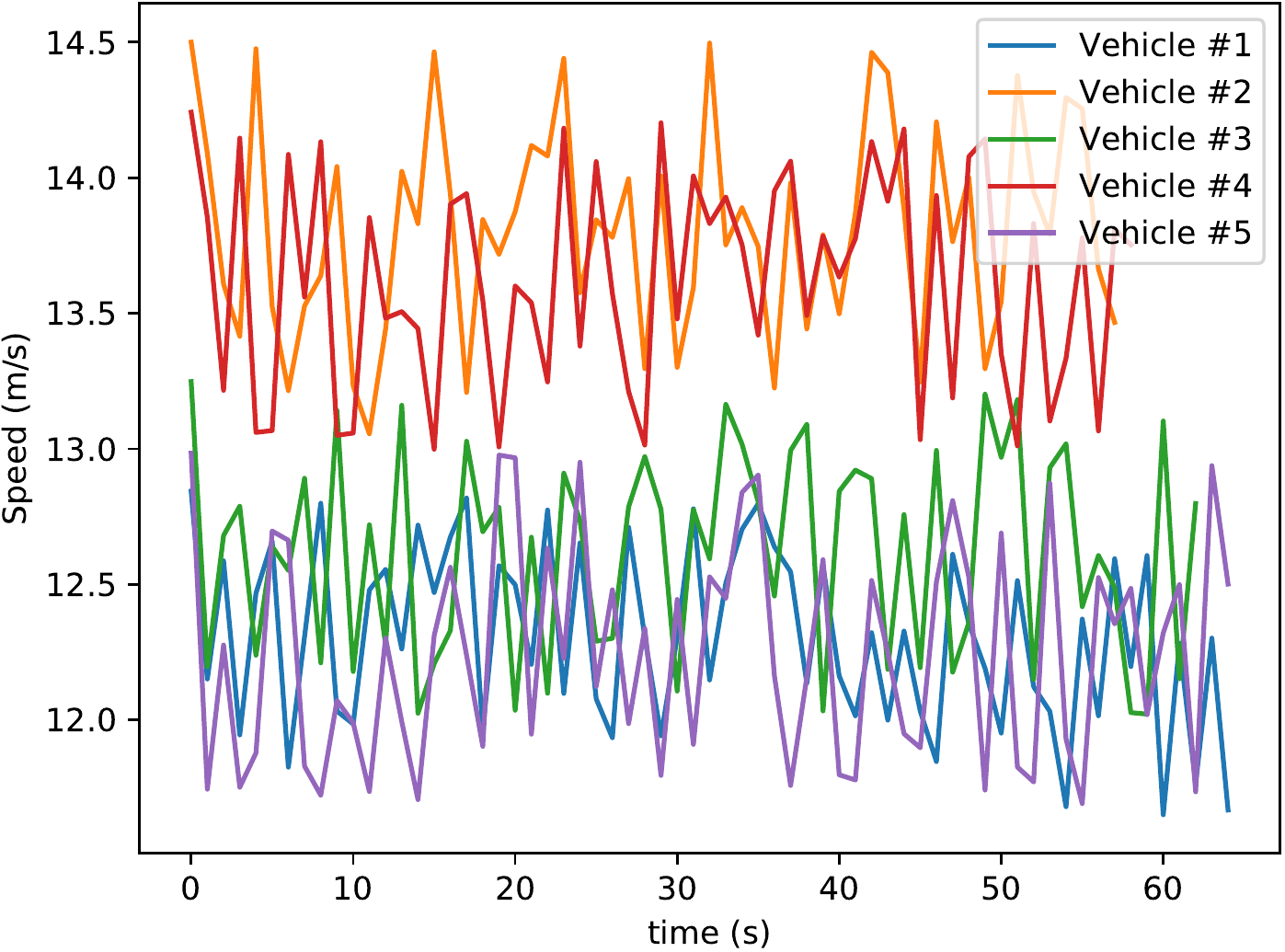}
        \caption{Constant speed limit.}
    \end{minipage}
    \begin{minipage}{0.6\linewidth}
        \includegraphics[width=\linewidth]{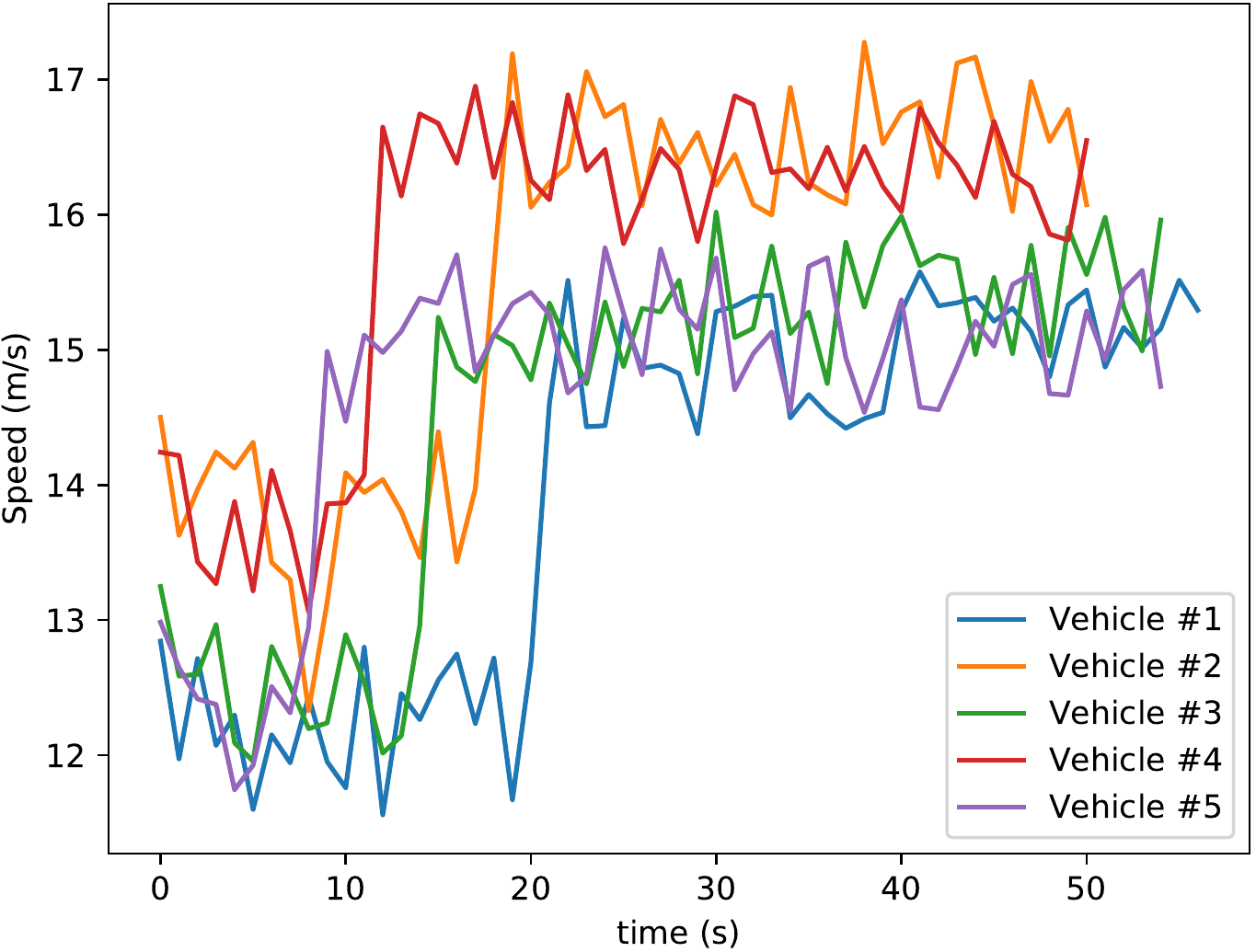}

        \caption{Speed limit raise.}
    \end{minipage}

    \begin{minipage}{1\linewidth}
        \vspace{0.6cm}
        \centering
        \includegraphics[width=0.55\linewidth]{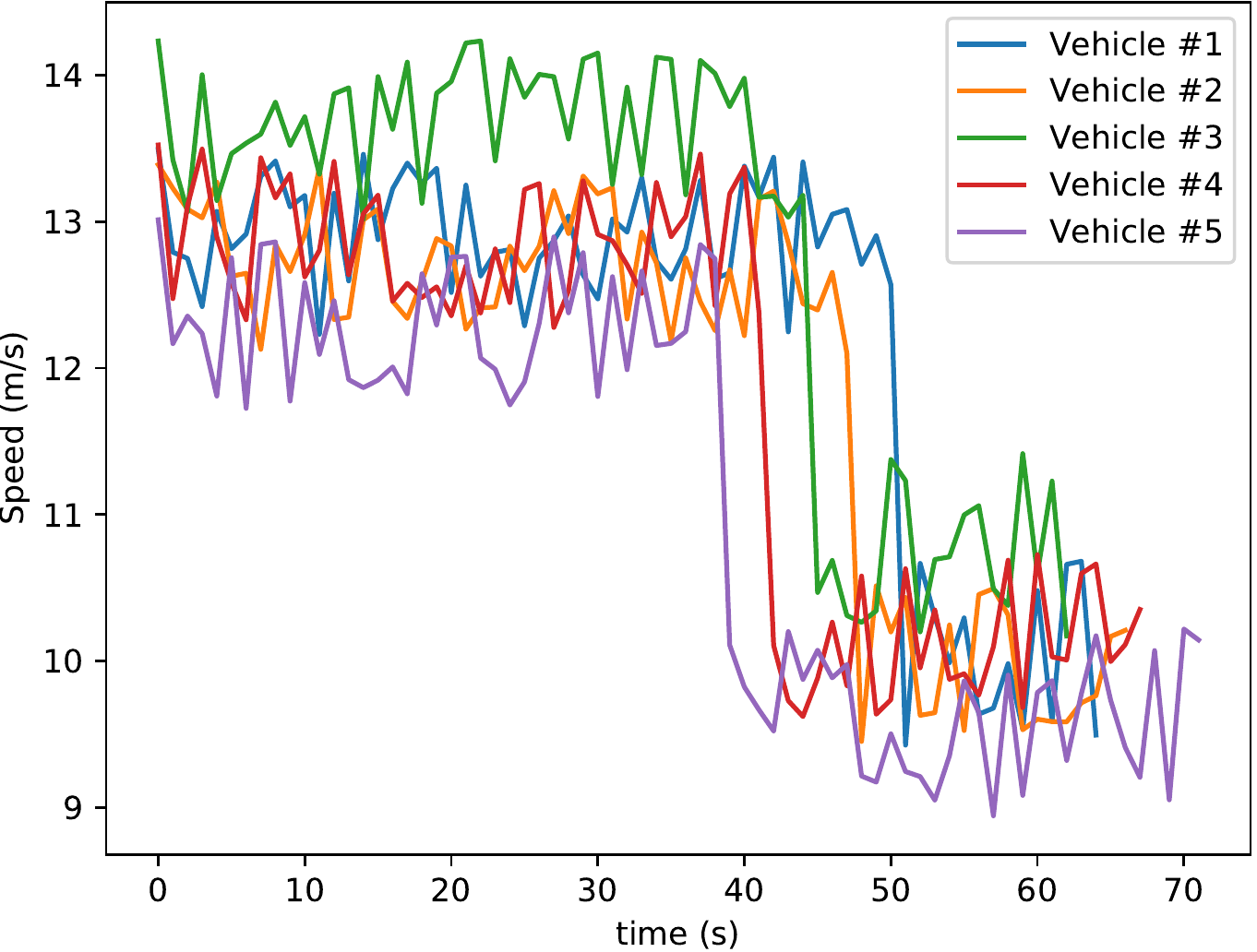}
        \caption{Speed limit decline.}
    \end{minipage}
    \caption{Driving scenarios simulated for the training dataset. }
    \label{speed}
\end{figure}

\subsection{Collected Data}
We generated two different datasets for the training and testing process.
The training dataset contains trajectory data of 6660 normal vehicles (across 1332 groups of cars) driving in a straight street.
The initial assigned speed of each vehicle is distributed randomly according to a Gaussian distribution (parameters are in \Table{distribution}), and it changes
based on one of the three scenarios in \Fig{speed}.

In the first scenario, we simulate a group of vehicles driving with a constant speed limit;
the second and third involve an increase and decrease of the speed limit in a road.
We use these scenarios to help our model learn
how a group of normal vehicles changes their behavior when it is necessary.

We also use simulated traffic in four streets (highlighted in \Fig{fig:calgary}) with different shapes, number of lanes and speed limits in the city to test our model.
Path \#1 is located in a residential area and contains one lane with a speed limit of \SI{40}{\km/\hour}.
Paths \#2 and \#3 are two main streets with two lanes, with maximum permitted driving speed of \SI{50}{\km/\hour}.
Also, vehicles can drive up to \SI{80}{\km/\hour} in the three lanes of path~\#3.

This dataset contains the trajectory data of $16320$ vehicles with
$1020$ abnormal cases.
Each abnormal case consists of a vehicle either over-speeding or under-speeding (as shown in \Table{distribution}).
Similar to normal cases, the speed of each abnormal case is chosen randomly.
\Table{distribution} shows the parameters of the Gaussian distribution used for each speed class.
The numbers in the table are multiplied by the speed limit of a street, for example the speed of a normal vehicle cannot exceed the street speed limit by more than $10\%$.

\begin{table}[ht]
    \centering
    \caption{Speed classes for each vehicle.}
    \label{distribution}
    \small
    \begin{tabularx}{8cm} {
        |>{\centering}p{\dimexpr.3\linewidth-2\tabcolsep-1.3333\arrayrulewidth} |>{\centering\arraybackslash}X |>{\centering\arraybackslash}X |>{\centering\arraybackslash}X |>{\centering\arraybackslash}X | }
        \hline
        SpeedClass  & $\sigma$ & $\mu$ & $\min$ & $\max$ \\
        \hline
        \multicolumn{5}{|c|}{$\times$ Speed Limit of streets} \\
        \hline
        Normal      & 1.0      & 0.1   & 0.9    & 1.1    \\
        Over Speed  & 1.25     & 0.1   & 1.2    & 1.3    \\
        Under Speed & 0.75     & 0.1   & 0.7    & 0.8    \\
        \hline
    \end{tabularx}
    \label{tab:table}
\end{table}

%% file: tex/evaluation.tex
\section{Evaluation Results}
\label{VI}

In this section, we evaluate the anomaly detection performance of \OurSolution\ and compare it with
three baseline methods: Dynamic Time Warping~(DTW)~\cite{R8}, Global Alignment Kernels (GAK)~\cite{R9}, and iForest~\cite{iforest}.

\subsection{Implemented Approaches}    DTW and GAK are two techniques to compare two or more time series with unequal length, which find an optimal alignment between the points on the two sequences.
We use a Python package called \textit{TsLearn}~\cite{Tslearn} which provides machine learning tools for the analysis of time series to implement these two methods.
Additionally, iForest is an unsupervised anomaly detection technique that works based on isolating the outlier cases.
We implement it using \textit{Sklearn}~\cite{scikit-learn}, an open source machine learning library which contains various tools for supervised and unsupervised learning.

Additionally, we use a Cosine similarity function instead of~\eqref{ex2} and study its performance compared to MSE.
To calculate the similarity of two latent vectors $A$ and $B$ using the Cosine similarity, we use the following expression:
\begin{equation}
    \textstyle
    \cos(A,B) = \frac{\sum_{n}^{i=1}A_i B_i}{\sqrt{\sum_{n}^{i=1}A_i^2} \sqrt{\sum_{n}^{i=1}B_i^2}}
\end{equation}

\begin{figure}[t]
    \centering
    \includegraphics[width=6cm]{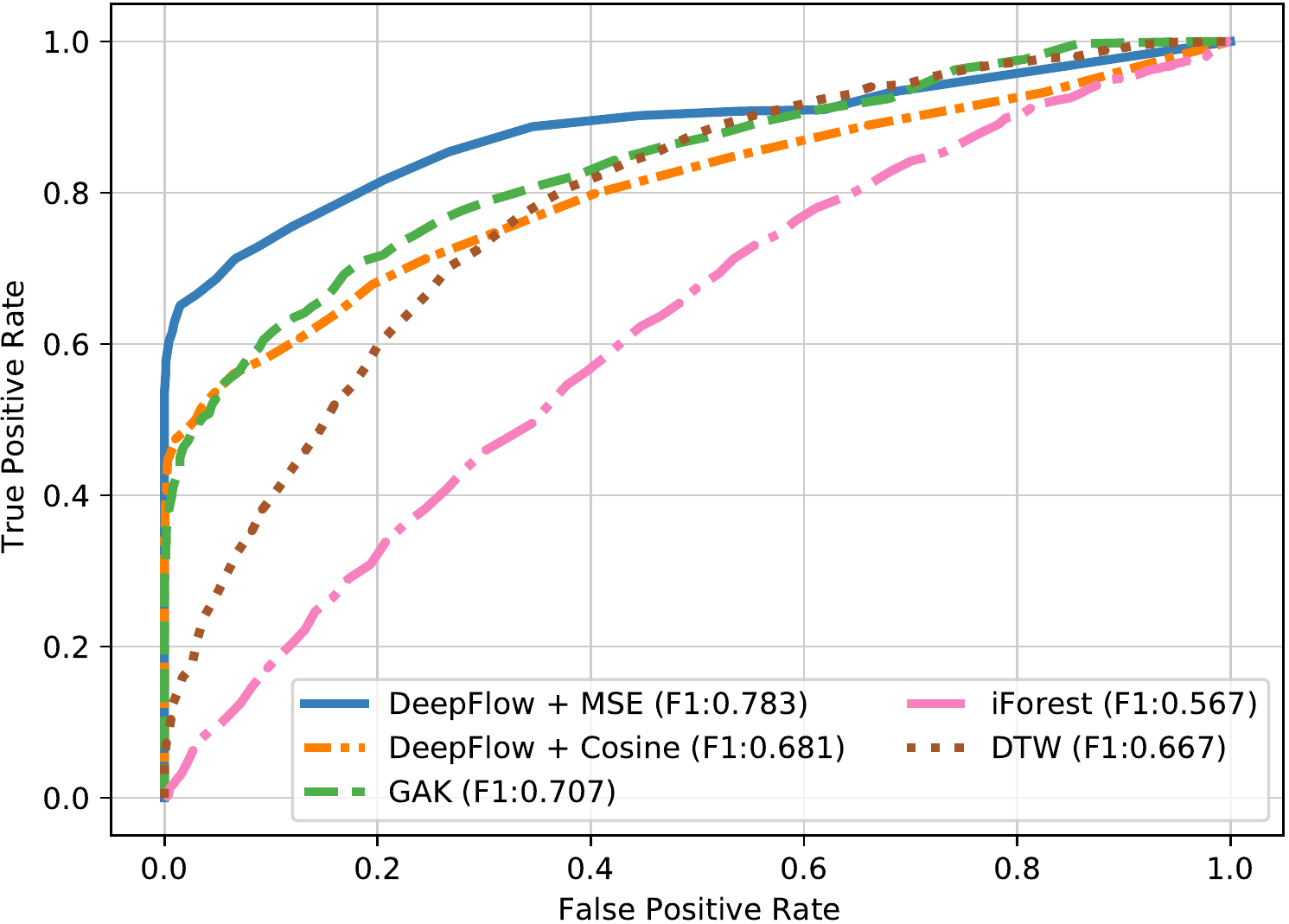}
    \caption{ROC curve for \OurSolution, DTW, GAK, and iForest.}\label{fig:roc}
\end{figure}

\begin{figure}[t]
    \centering
    \includegraphics[width=6cm]{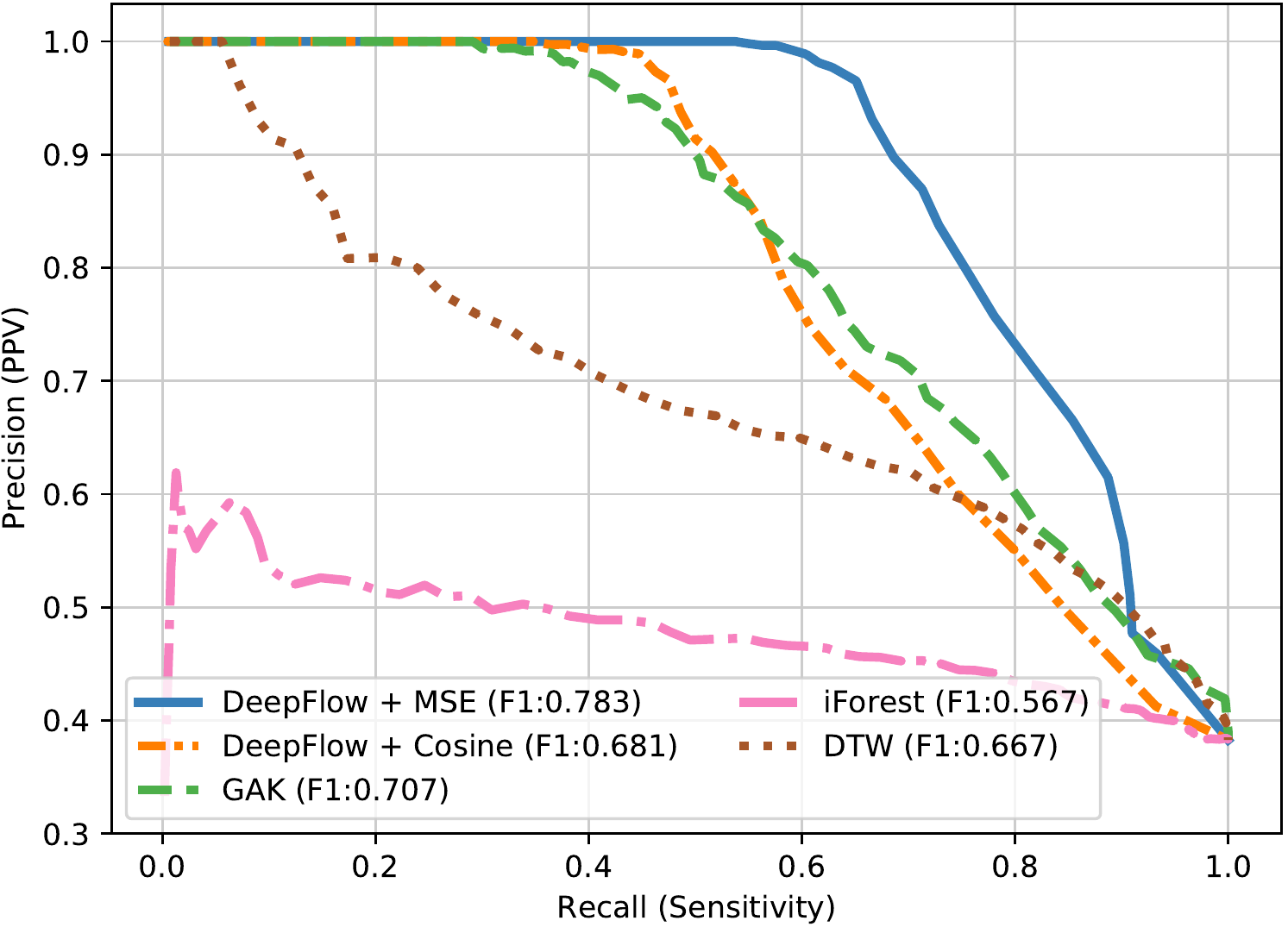}
    \caption{Precision-Recall curve for \OurSolution, DTW, GAK, and iForest. \vspace{-0.5cm}}\label{fig:pr}
\end{figure}

\begin{table}[t]
    \small
    \centering
    \caption{Performance evaluation.}
    \label{result}
    \hspace*{-0.5cm}
    \begin{tabularx}{10cm} {
        |>{\centering}p{\dimexpr.34\linewidth-2\tabcolsep-1.3333\arrayrulewidth}|>{\centering\arraybackslash}X |>{\centering\arraybackslash}X |>{\centering\arraybackslash}X |>{\centering\arraybackslash}X | }
        \hline
        Method                       & Recall           & Precision        & F1               & Time                   \\
        \hline
        \textbf{\OurSolution\ (MSE)} & 71.27\%          & \textbf{86.96\%} & \textbf{78.34\%} & \multirow{2}{*}{96$ms$} \\
        \OurSolution\ (Cosine)       & 67.84\%          & 68.38\%          & 68.11\%          &                          \\
        GAK                          & 70.88\%          & 70.54\%          & 70.71\%          & \textbf{33$ms$}         \\
        DTW                          & 78.92\%          & 41.40\%          & 66.75\%          & 136$ms$                 \\
        iForest                      & \textbf{89.90\%} & 57.83\%          & 56.69\%          & 154$ms$                 \\
        \hline
    \end{tabularx}
\vspace{-0.5cm}
\end{table}

\subsection{Performance Metrics}
Undetected abnormal vehicles in the system are a threat to traffic safety;
on the other hand, marking normal vehicles as abnormal is not desirable either.
Therefore, finding a proper threshold has a significant impact on the True Positive and False Positive rate of the system.
We employ a \textbf{Receiver Operating Characteristic (ROC)} curve to show \OurSolution's performance compared to other solutions;
this displays the True Positive rate~(TPR) versus False Negative rate~(FNR)
when we change the anomaly score threshold.
TPR shows the proportion of abnormal cases which are detected, and FNR indicates the proportion of normal cases which are marked incorrectly.

We also use the \textbf{Precision-Recall} metric to evaluate the quality of our detector.
Precision-Recall is an important measure when we use an imbalanced dataset.
Precision is defined as the ratio of between the number of true positives and the number of detected anomalies, while
recall determines the proportion of the actual abnormal cases that are identified correctly.
A high value in both metrics is a sign of good performance for the detector.
To consider Precision and Recall metrics in our evaluations,
we use $F_{1}$ Score as it places equal weights on both precision and recall;
this can be expressed as follows:
\vspace{0.2cm}
\begin{equation}
    \textstyle
    F_1 = 2 \times \frac{precision \times recall}{ precision + recall}\label{eq:F1}
\end{equation}

\subsection{Results and Discussion}

\cat{Performance Comparison}
We use the dataset that is described in \Section{V}.
The results, shown in \Fig{fig:roc}, \Fig{fig:pr} and \Table{result}, indicate that \OurSolution\ (with MSE) outperforms other solutions.
%

\begin{figure}[t]
    \centering
    \includegraphics[width=9cm]{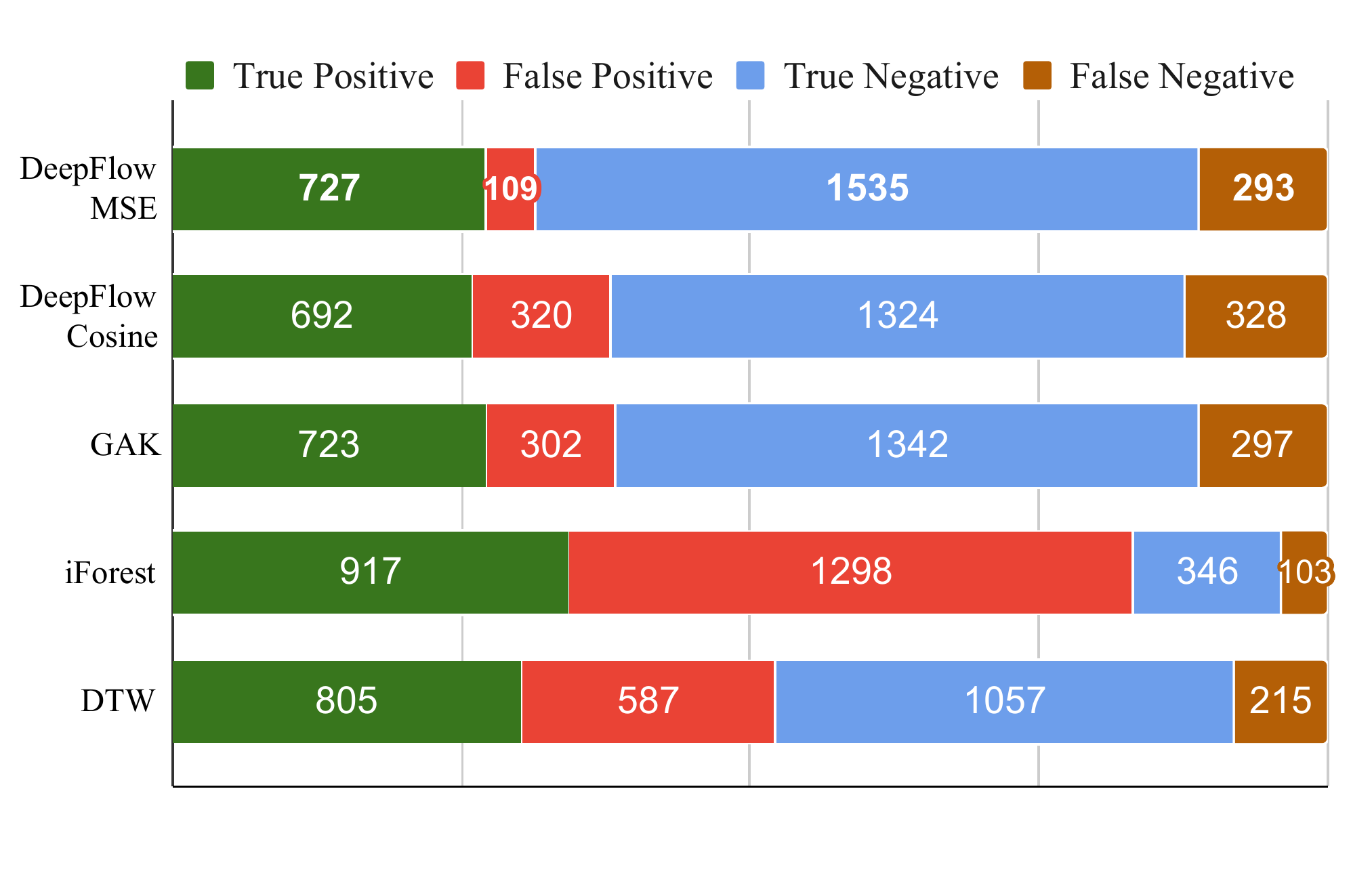}
    \vspace{-1cm}
    \caption{TP, FP, TN, and FN for \OurSolution, DTW, GAK,
        and iForest.}
    \label{TP}
\end{figure}
Based on \Fig{fig:roc}, there is a threshold where \OurSolution\ detects $70\%$ of the abnormal cases or a $6\%$ FN rate.
However, achieving the same TP rate using GAK or DTW will result in an FN rate of 18\% and 26\%, respectively.
Likewise, the Precision-Recall curve in \Fig{fig:pr} shows that \OurSolution\  can achieve both greater precision and recall than other methods.
In addition, \Table{result} shows that \OurSolution\ can reach the highest $F_{1}$ score among all the evaluated approaches.
Also, we show the average anomaly detection time for each traffic flow in the table.

\cat{Microscopic Behavior}
As \Fig{TP} shows, \OurSolution\ achieved the lowest number of false positives and detected more than 700 abnormal cases.
Although iForest and DTW detected more abnormal cases than \OurSolution, a large number of normal cases are incorrectly marked by these approaches.
The reason is that the random distribution of the vehicles speeds makes it difficult for these classifiers to distinguish between the normal and abnormal cases, thus resulting in
high false positive rates.
However, \OurSolution\ learns the pattern of similarity between the normal behaviors, which help it detect anomalies more accurately.

\cat{Anomaly Threshold}
Finding a proper threshold value can affect the accuracy of the system noticeably.
This value can be selected for each street individually or a common value can be used.
\Table{tab:streets} shows that finding a threshold specifically for each street increases the $F_{1}$ score.
However, it requires providing a dataset containing abnormal cases for each path and calculating the corresponding threshold value.
Therefore, we suggest using a common threshold for the majority of the paths in cities and only assign individual thresholds
for streets with greater importance, \eg\ highways.

\cat{Road Characteristics}
From \Table{tab:streets}, we can observe that the performance of \OurSolution\ is dependent on the shape and
length of the street that is being monitored.
For example, street \#2 is the longest simulated path which results in our model being unable to detect abnormal cases as accurately as in other paths.
Also, street \#3 has the most similar structure to the training dataset among other paths, so
our model is able to detect anomalies with a higher $F_{1}$ score on this path.

\begin{table}[t]
    \small
    \centering
    \vspace{0.5cm}
    \caption{\OurSolution\ $F_{1}$ score for streets shown in \Fig{fig:calgary}.}
    \begin{tabularx}{8cm} {|>{\centering}p{\dimexpr.1\linewidth-2\tabcolsep-1.3333\arrayrulewidth} |>{\centering}p{\dimexpr.15\linewidth-2\tabcolsep-1.3333\arrayrulewidth} |>
            {\centering}p{\dimexpr.15\linewidth-2\tabcolsep-1.3333\arrayrulewidth} |>{\centering\arraybackslash}X |}
        \hline
        Path
        & Indiv.
        & Com.
        & Feature \\
        \hline
        1 & 81.4\% & 74.9\% & Includes turnings              \\
        2 & 82.5\% & 60.0\% & Curved and longest             \\
        3 & 95.4\% & 86.4\% & Straight (similar to training) \\
        4 & 78.0\% & 76.4\% & North to south                 \\
        \hline
    \end{tabularx}
    \label{tab:streets}
    \vspace{-0.5cm}
\end{table}

\cat{Parameter Tuning}
We analyze the result of changing the value of two different training parameters on our model behavior.
First, we investigate the effect of the $\lambda$ value in \Exp{loss} on the training performance.
The results show that best performance is when the importance of the two loss functions in the training phase are equal,
which can be achieved by setting the $\lambda$ value to 1.
Second, we searched for the best latent vector size.
Our experiment shows that we can, on average, get the best result from compressing the input value by 60\%.

%% file: tex/conclusion.tex
\section{Conclusion}
\label{VII}

In this paper, we presented \OurSolution\ for detecting abnormal traffic flows.
We showed that \OurSolution\ performs well even when it is trained with a dataset that is collected in a different environment.
This feature indicates that \OurSolution\ can be employed for roads with different shapes, number of lanes and speed limits, with no retraining required.
Even though that \OurSolution\ can work with any number of inputs (\eg\ speed, acceleration and steering angle), we used trajectory data to detect anomalies as
this type of data can be collected easily using existing surveillance camera infrastructure at no additional cost.
%
Despite the excellent performance and easy implementation, our model suffers from two disadvantages.
First, this model only works for a group of vehicles and cannot detect abnormal behavior of an individual car.
The second problem is, we assume abnormal vehicles form the minority of target traffic flow, so our model only works
as long as this assumption holds.

Future works include determining the effect of using more variables (such as speed, acceleration, and vehicle angle) and considering a wider range of anomalies.